\documentclass[10pt,twocolumn,letterpaper]{article}

\usepackage{wacv}
\usepackage{hyperref}
\usepackage{times}
\usepackage{epsfig}
\usepackage{graphicx}
\usepackage{amsmath}
\usepackage{amssymb}
\usepackage{subcaption}

\graphicspath{ {images/} }

\wacvfinalcopy % *** Uncomment this line for the final submission

\def\wacvPaperID{58} % *** Enter the wacv Paper ID here

\ifwacvfinal\fi
\begin{document}

%%%%%%%%% TITLE
\title{Leveraging Uncertainty Estimates for Predicting Segmentation Quality}

% Authors at the same institution
%\author{First Author \hspace{2cm} Second Author \\
% Authors at different institutions

% \author{First Author \\
% \and
% Second Author \\

\author{Terrance DeVries \\
University of Guelph and Vector Institute\\
{\tt\small terrance@uoguelph.ca}
\and
Graham W. Taylor \\
University of Guelph and Vector Institute\\
Canadian Institute for Advanced Research\\
{\tt\small gwtaylor@uoguelph.ca}
}

\maketitle
\ifwacvfinal\thispagestyle{empty}\fi

%%%%%%%%% ABSTRACT
\begin{abstract}
The use of deep learning for medical imaging has seen
tremendous growth in the research community. One reason for the slow uptake of
these systems in the clinical setting is that they are complex, opaque and tend
to fail silently. Outside of the medical imaging domain, the machine learning
community has recently proposed several techniques for quantifying model
uncertainty (i.e.~a model knowing when it has failed). This is  important in
practical settings, as we can refer such cases to manual inspection or
correction by humans. In this paper, we aim to bring these recent results on
estimating uncertainty to bear on two important outputs in deep learning-based
segmentation. The first is producing spatial uncertainty maps, from which a
clinician can observe where and why a system thinks it is failing.
The second is quantifying an image-level prediction of failure, which is
useful for isolating specific cases and removing them from automated pipelines.
We also show that reasoning about spatial uncertainty, the first output, is a
useful intermediate representation for generating segmentation quality
predictions, the second output. We propose a two-stage architecture for producing these
measures of uncertainty, which can accommodate any deep learning-based medical
segmentation pipeline.
\end{abstract}

%%%%%%%%% BODY TEXT
\section{Introduction}

In recent years, the use of deep learning for medical imaging tasks has increased in prevalence, with these powerful algorithms being applied to a wide variety of medical imaging applications, from metastasis detection for breast cancer \cite{Liu2017-xi}, to improving reconstruction for medical resonance imaging \cite{Mardani2017-ft}. In some cases, deep learning has even matched or exceeded human performance, such as on the tasks of skin lesion classification \cite{esteva2017dermatologist} and identifying diabetic retinopathy \cite{gulshan2016development}.

Unfortunately, despite the recent successes reported in the literature, we have yet to see the widespread adoption of deep learning in clinical settings. One possible reason for this delay could be the lack of suitable uncertainty estimates \cite{litjens2017survey}. Current neural network-based models are often incapable of indicating when their predictions may be faulty, and as a result they fail silently, without any indication that a mistake has been made. This behaviour is worrying for applications that rely on accurate uncertainty estimates for decision making, such as those that a medical professional might encounter when diagnosing a patient based on the results of a predictive model. Proper uncertainty estimates would allow those reviewing the predictions to act accordingly in order to prevent undesirable outcomes \cite{amodei2016concrete}.

One area where this is a problem is the task of segmenting medical images. If the result of an automated segmentation is poor, we would like to refer the case to a qualified human for follow-up. Furthermore, given the spatial nature of the task, it would be useful to provide the human with a map of the model's uncertainty so that they can better understand where and why the model failed, and perhaps take the uncertainty estimates into consideration when manually correcting the segmentation.

In this work, we consider the specific task of segmenting skin lesions. However, we propose a framework that is general enough to support a variety of medical segmentation tasks. We propose learning spatial uncertainty maps for each segmentation, which can then be used to improve our prediction of the quality of the segmentation, and we demonstrate that this yields an improved performance over alternative techniques that use deep learning to predict segmentation quality. Based on our finding that per-pixel uncertainty is a useful intermediate representation for predicting image-level segmentation quality we compare several contemporary uncertainty estimation methods to assess their relative merits.

The key contribution of this work is unifying two pursuits that have to-date remained disparate: uncertainty estimation in deep neural networks and predicting image-level segmentation quality. Our method is easy to deploy in practice, as it is modular and agnostic to the specific deep-learning architecture or uncertainty estimation technique. Secondary contributions are the extension of Learning Confidence Estimates~\cite{devries2018learning} to pixel-level, rather than scalar output and the empirical finding that at least three recently-proposed methods for quantifying uncertainty can aid almost equally well in predicting segmentation quality.

\section{Related Work}
Our work attempts to unite two research areas that are of interest to the machine learning and computer vision community: uncertainty estimation and predicting segmentation quality. Here we provide a brief overview of relevant recent work in the respective areas.

\subsection{Uncertainty Estimation}
Uncertainty estimates are useful in the context of deployed machine learning systems as they have been shown to be capable of detecting when a neural network is likely to make an incorrect prediction, or when an input may be out-of-distribution.

Traditionally, much of the work done on uncertainty estimation techniques is inspired by Bayesian statistics. A classic example is the Bayesian Neural Network (BNN) \cite{neal2012bayesian}, which attempts to learn a distribution over each of the network's weight parameters. Such a network would be able to produce a distribution over the output for any given input, thereby naturally producing uncertainty estimates. Unfortunately, Bayesian inference is computationally intractable for these models in practice, so much effort has been put into developing approximations of Bayesian neural networks that are easier to train.

Recent efforts in this area include Monte-Carlo Dropout \cite{gal2016dropout}, Multiplicative Normalizing Flows \cite{louizos2017multiplicative}, and Stochastic Batch Normalization \cite{atanov2018uncertainty}. These methods have been shown to be capable of producing uncertainty estimates, although with varying degrees of success. The main disadvantage with these BNN approximations is that they require sampling in order to generate the output distributions. As such, uncertainty estimates are often time-consuming or resource-intensive to produce, often requiring 10 to 100 forward passes through a neural network in order to produce useful uncertainty estimates at inference time.

An alternative to BNNs is Deep Ensembles \cite{lakshminarayanan2017simple}, which proposes a frequentist approach to the problem of uncertainty estimation by training many models and observing the variance in their predictions. However, this technique is still quite resource intensive, as it requires inference from multiple models in order to produce the uncertainty estimate.

A promising alternative to sampling-based methods is to instead have the neural network \textit{learn} what its uncertainty should be for any given input, as demonstrated in \cite{kendall2017uncertainties} and \cite{devries2018learning}. These methods are more computationally efficient compared to other techniques, and thus better suited when computational resources are limited or when real-time inference is required.

\begin{figure*}[t]
\centering
\includegraphics[width=\textwidth, trim={0 5cm 0 5cm}, clip]{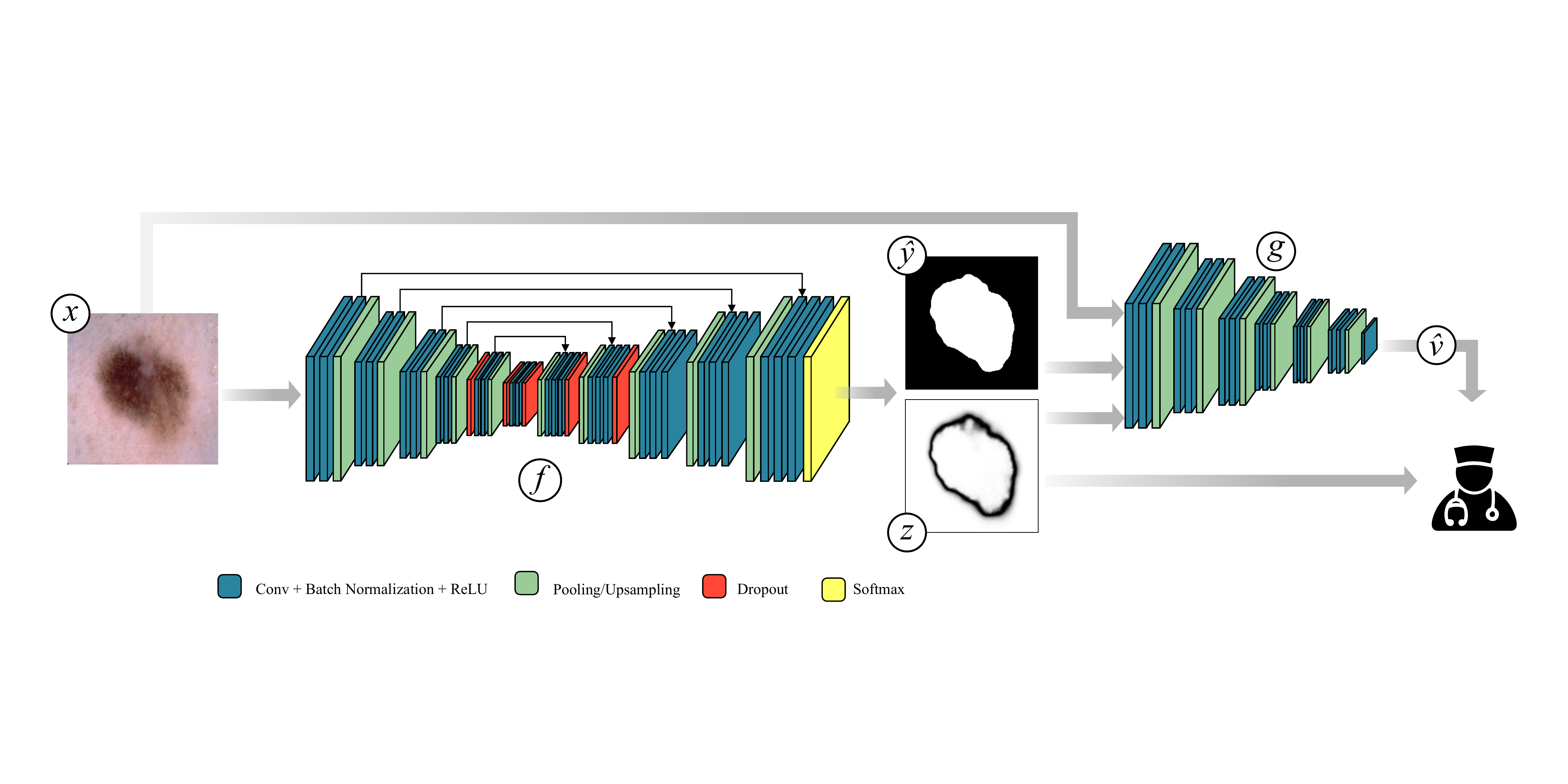}
\caption{System diagram for our proposed pipeline. A semantic segmentation network $f$ takes an input image $x$ and produces a segmentation prediction $\hat{y}$ and an uncertainty map $z$. A segmentation quality network $g$ then receives as input $x$, $\hat{y}$, and $z$ to produce a quality estimate $\hat{v}$. The uncertainty map can be used to interpret the segmentation network's output, while the quality estimate can be used to automatically reject, or alert the user to poor segmentations. This diagram depict a system which utilizes a maximum softmax probability uncertainty map. Other uncertainty estimation methods may have small architectural differences.}
\label{fig:system_diagram}
\end{figure*}

\subsection{Segmentation Quality Prediction}
When applying uncertainty estimates to the task of semantic segmentation, a number of works have proposed ways to produce spatial \emph{uncertainty maps}, which visualize a model's confidence in its predictions for each pixel in the image. In most cases, uncertain regions are likely to be misclassified, and the uncertainty maps allow one to see which parts of the image are likely to be problematic \cite{kampffmeyer2016semantic, kendall2015bayesian, kendall2017uncertainties}. This feature gives the model some amount of interpretability, and provides the end user with more information with which they can decide whether the final segmentation is to be trusted or how it should be modified (e.g.~in a human-in-the-loop setting).

However, if the semantic segmentation model is part of a larger automated pipeline, pixel-level uncertainty estimates are not as useful, as perfectly acceptable segmentations can still contain some uncertainty. In this case, it is more useful to create a model that can predict the quality of the segmentation at the whole-image level. Previous efforts have attempted to learn the quality of segmentations from hand-crafted image or segmentation features \cite{kohlberger2012evaluating, zhang2006meta}, but these approaches are limited by the expressiveness of their respective hand-crafted features. They are also limited in their transferability across different medical imaging modalities.

Contemporary approaches have exploited the powerful feature learning capabilities of deep learning. Recently, adversarial training has been used to improve the performance of convolutional segmentation networks by having an auxiliary discriminator network predict the quality of the segmentation (i.e.~whether or not the predicted segmentation is discernible from a ground truth segmentation)~\cite{luc2016semantic}. The segmentation network then uses this information to improve its predictions and produce more realistic looking segmentations. This technique has previously been demonstrated to improve segmentation quality in medical imaging tasks such as prostate cancer or brain MRI segmentation \cite{kohl2017adversarial,moeskops2017adversarial}. Adversarial training works well to improve segmentation quality, but the quality estimation network has limited utility as the outputs don't have any human interpretable meaning associated with them beyond whether the segmentation looks realistic or not.
% Not sure if this whole interpretability argument is convincing

As a solution to the interpretability issue, methods have been proposed which attempt to predict segmentation quality in terms of metrics that are more meaningful to humans, such as Jaccard index or Dice coefficient. An example of this is QualityNet \cite{huang2016qualitynet}, which learns a direct mapping between a masked input image and its corresponding segmentation quality via a convolutional neural network (CNN). Another interesting approach is Reverse Classification Accuracy (RCA) \cite{valindria2017reverse}, which evaluates segmentation quality by training a reverse classifier on a predicted segmentation from a new image, and then evaluating on a set of reference images that have ground truths available. While these techniques work well in their respective settings, none of them exploit uncertainty information, which could be used to improve the accuracy of the segmentation quality prediction.

\section{Estimating Segmentation Quality with Uncertainty Information}\

In order to leverage uncertainty information in our predictions of segmentation quality, we first train a neural network-based semantic segmentation model $f$ (as shown in Figure~\ref{fig:system_diagram}). The segmentation network takes in some image $x$, and produces two outputs: class prediction logits $\rho$ and corresponding uncertainty (or confidence) estimates $z$:
\begin{equation}
\rho, z = f(x) .\
\end{equation}
As uncertainty is estimated per pixel, we refer to $z$ as an uncertainty map. In this formulation, $z$ may either be calculated using the original outputs of $f$, or the network can produce the $z$ directly. To obtain the final segmentation prediction $\hat{y}$ we take the argmax of the prediction logits or the class prediction probabilities:
\begin{equation}
\hat{y} = \text{argmax}(\rho) .\
\end{equation}
%In our experiments we use a U-Net architecture \citep{ronneberger2015u} for semantic segmentation, but our approach should generalize to any neural network based segmentation architecture. To produce the uncertainty estimates we experiment with several different methods.
A second network $g$ is then trained to predict the quality of the segmentation $\hat{v}$, given the original input image $x$, as well as the  predicted segmentation mask $\hat{y}$ and uncertainty map $z$ from $f$:
\begin{align}
\hat{v} &= g(x, \hat{y}, z) .\
\end{align}
Under our framework, the segmentation quality measurement can be any segmentation-based evaluation metric, or even multiple metrics predicted simultaneously. To obtain the true segmentation quality labels $v$ to train $g$, we evaluate the segmentation predictions from $f$ using the ground truths from the training set. The training set for $g$ can be the same one as used to train $f$, or a separate holdout set, or a combination of the two. In the case that $f$ performs very well on the training set, a holdout set may be necessary, as the lack of examples of poor segmentations will bias $g$ towards always predicting that the segmentation is good.

There have been many methods proposed for attaining uncertainty or confidence estimates from neural networks, but for our experiments we consider four  methods: maximum softmax probability, Monte-Carlo Dropout, heteroscedastic classifier neural networks, and learned confidence estimates. We selected these based on their simplicity to implement as well as their diversity.

\subsection{Maximum Softmax Probability}
The first method we evaluate is the maximum softmax probability, which was demonstrated by \cite{hendrycks2016baseline} to be surprisingly effective at the tasks of misclassification and out-of-distribution detection. The softmax probability can be obtained from any classification neural network for free, making it an appealing choice for confidence estimates. To calculate the maximum softmax probability we simply calculate the maximum across the class dimension of the softmax output from the network $f$:
\begin{align}
z = \text{max}(\text{Softmax}(\rho)) .\
\end{align}
For segmentation, this is done per output pixel in order to obtain an uncertainty map that is of the same resolution as the input image.

\subsection{Monte-Carlo Dropout}
The second uncertainty estimation method we consider is Monte-Carlo dropout (MC-dropout) \cite{gal2016dropout}, which has previously seen success in the field of medical imaging \cite{leibig2017leveraging, yang2016fast}. MC-dropout approximates a BNN by sampling from a neural network trained with dropout \cite{srivastava2014dropout} at inference time in order to produce a distribution over the outputs. This approach is very simple to implement in practice, and as many modern neural network architectures already leverage dropout for regularization purposes, uncertainty estimates can often be attained without any changes to the architecture or training paradigm. MC-dropout models epistemic uncertainty, which is the uncertainty associated with the model parameters, such that increasing the amount of training data tends to decrease the epistemic uncertainty associated with the model.

Following the approach used for Bayesian SegNet \cite{kendall2015bayesian,kendall2017uncertainties}, we apply dropout with $p=0.5$ after each central convolutional block of our U-Net architecture. During test time we sample from the segmentation network $T$ times (we use $T = 20$) and then calculate the average softmax probability over all of the samples in order to approximate Monte Carlo integration:
\begin{equation}
p = \frac{1}{T} \sum_{t=1}^{T} \text{Softmax}(\rho_{t}) .\
\end{equation}
Model uncertainty $z$ is estimated by calculating the entropy of the averaged probability vector across the class dimension:
\begin{equation}
z= -\sum_{c=1}^{C} p_{c} \text{log} p_{c}.\
\end{equation}

\subsection{Heteroscedastic Classifier Neural Network}
The third uncertainty estimation technique we evaluate is one which attempts to model aleatoric uncertainty, which is the uncertainty present in the data itself, such as from noisy labels or measurements. To model aleatoric uncertainty, \cite{kendall2017uncertainties} introduce the heteroscedastic classifier neural network, which we will refer to as HCNN. In this method, uncertainty estimates are learned by the network, rather than being calculated post-hoc as with MC-dropout. The HCNN produces two outputs via two separate output branches: class prediction logits, and a variance estimate which represents model uncertainty. Again, in the case of segmentation, these two quantities are computed per output pixel. During training, Gaussian noise with magnitude equal to the variance estimate is sampled and added to the probability logits, which are used to calculate the training loss as usual:
\begin{equation}
p=\frac{1}{T} \sum_{t=1}^{T}\text{Softmax}(\rho+z \epsilon_{t}), \quad \epsilon_{t} \sim \mathcal{N}(0, 1) .\
\end{equation}
In our experiments we set T = 100. We also apply a softplus function to the output of the variance estimation branch in order to ensure that it is non-negative.

\subsection{Learned Confidence Estimates}
The final technique we evaluate is Learned Confidence Estimates (LCE), which was introduced by \cite{devries2018learning}. This method is similar to HCNN in that the network produces two separate outputs: prediction probabilities and a confidence estimate. Confidence estimates are motivated by interpolating between the predicted probability distribution and the target distribution during training, where the degree of interpolation is proportional to the confidence estimate:
\begin{equation}
p = z \cdot \text{Softmax}(\rho) + (1-z) \cdot \text{Onehot}(y) .\
\end{equation}
In this formulation, low confidence estimates are pushed towards the correct answer, while high confidence estimates remain unchanged. To prevent the model from always producing low confidence estimates, a log penalty on the confidence estimate is added to the loss function. As a result, the network can reduce its overall training loss if it correctly infers which samples it is likely to predict incorrectly.

\begin{figure*}[t]
\centering
\includegraphics[width=0.76\textwidth]{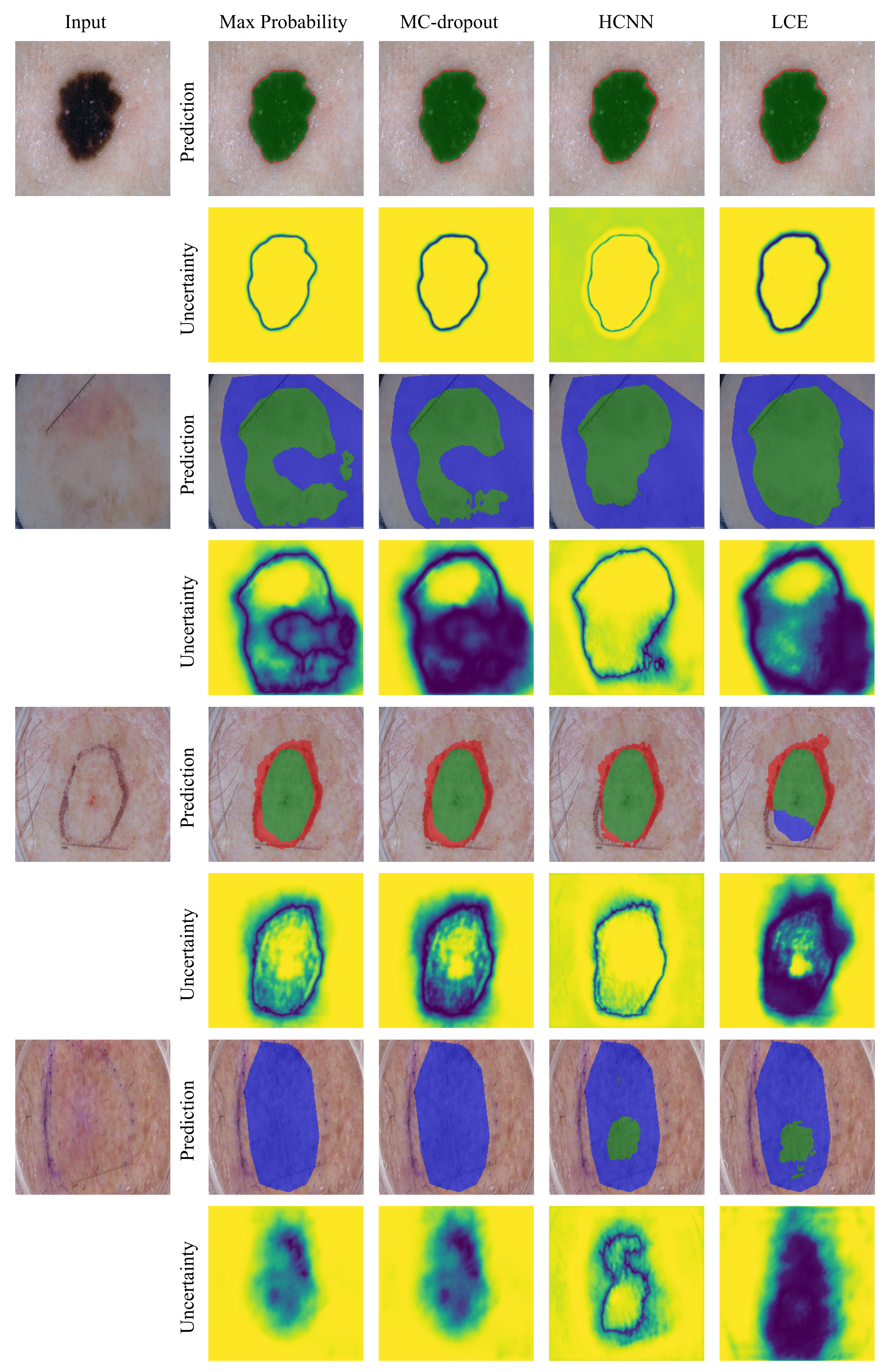}
\caption{Segmentation predictions (even rows) and uncertainty maps (odd rows) for four different uncertainty estimation methods. In the segmentation predictions, green = true positive, red = false positive, and blue = false negative. In the uncertainty maps yellow = low uncertainty and purple = high uncertainty. Best viewed in colour.}
\label{fig:uncertainty_maps}
\end{figure*}

\begin{table*}[t]
\centering
\caption{Comparison of different Jaccard index estimation methods. $\downarrow$ indicates that a lower score is better, while $\uparrow$ indicates that a higher score is better. All values except for RMSE are percentages. Results averaged over 5 runs with parameters randomly initialized.}
\label{table:seg_quality_prediction}
\begin{tabular}{lccccc}
\hline
\textbf{Method} & \multicolumn{1}{c}{\textbf{\begin{tabular}[c]{@{}c@{}}RMSE\\ \\ $\downarrow$ \end{tabular}}} & \textbf{\begin{tabular}[c]{@{}c@{}}Detection\\ Error\\ $\downarrow$ \end{tabular}} & \textbf{\begin{tabular}[c]{@{}c@{}}AUROC\\ \\ $\uparrow$ \end{tabular}} & \textbf{\begin{tabular}[c]{@{}c@{}}AUPR-\\ Pass\\ $\uparrow$ \end{tabular}} & \textbf{\begin{tabular}[c]{@{}c@{}}AUPR-\\ Fail\\ $\uparrow$ \end{tabular}} \\
\hline
RCA & 0.438 $\pm$ 0.007 & 43.8 $\pm$ 1.0 & 53.7 $\pm$ 1.4 & 74.4 $\pm$ 1.4 & 30.7 $\pm$ 0.9 \\
QualityNet & 0.213 $\pm$ 0.009 & 25.7 $\pm$ 2.7 & 80.9 $\pm$ 3.1 & 89.0 $\pm$ 2.3 & 69.1 $\pm$ 4.7 \\
\hline
No Uncertainty & 0.198 $\pm$ 0.011 & 27.3 $\pm$ 3.3 & 79.8 $\pm$ 3.8 & 88.5 $\pm$ 1.9 & 66.4 $\pm$ 7.9 \\
Max Probability & \textbf{0.168} $\pm$ 0.014 & \textbf{18.4} $\pm$ 3.0 & \textbf{88.4} $\pm$ 2.2 & \textbf{93.2} $\pm$ 1.6 & \textbf{80.5} $\pm$ 3.2 \\
MC-dropout & \textbf{0.163} $\pm$ 0.010 & \textbf{18.8} $\pm$ 1.4 & \textbf{88.1} $\pm$ 0.8 & \textbf{93.5} $\pm$ 1.3 & \textbf{78.1} $\pm$ 3.0 \\
HCNN & 0.196 $\pm$ 0.023 & 21.3 $\pm$ 1.8 & 85.5 $\pm$ 1.5 & 91.6 $\pm$ 1.4 & 76.2 $\pm$ 4.5 \\
LCE & \textbf{0.167} $\pm$ 0.019 & \textbf{19.3} $\pm$ 1.1 & \textbf{88.3} $\pm$ 1.4 & \textbf{93.6} $\pm$ 1.5 & \textbf{79.1} $\pm$ 3.9 \\
\hline
\end{tabular}
\end{table*}

\section{Experiments}
To evaluate our method we apply it to the problem of skin lesion segmentation, as this application has received a fair amount of attention from the deep learning community \cite{esteva2017dermatologist}. Specifically, we work with the ISIC 2017 dataset \cite{codella2017skin}, which consists of 2,750 dermoscopic images in three official dataset splits: 2,000 training images, 150 validation images, and 600 test images. Each image depicts a skin lesion from one of three different classes: melanoma, seborrheic keratosis, and benign nevi. Additionally, each image has an accompanying expert-labeled binary segmentation mask. For our experiments, we resize all images and ground truth masks to $224 \times 224$ pixels.

For our semantic segmentation network $f$, we adopt a U-Net style model architecture \cite{ronneberger2015u}. To facilitate MC-dropout we apply dropout with $p=0.5$ to the central layers of the encoder and decoder, as in Bayesian SegNet \cite{kendall2015bayesian}. Each model is trained for 120 epochs using batches of 16 images, and the Adam optimizer \cite{kingma2014adam} with a learning rate of 0.001. Images are randomly flipped and rotated at 90 degree intervals for data augmentation. For each uncertainty estimation method we train five models with random parameter initializations so that we can observe variance in performance. We find that all segmentation networks score within the range of $0.73 \pm 0.02$ Jaccard index, which is competitive with single-model performance for this task.

Our segmentation quality prediction network $g$ is a VGG-style CNN, which is trained to predict the Jaccard index of any given segmentation prediction given the original image, predicted segmentation mask, and confidence. We train our quality prediction network for 30 epochs using batches of 16 images, and the Adam optimizer with a learning rate of 0.001. As with our segmentation network, we apply flipping and rotating transforms for data augmentation.

For comparison, we also train models using two other neural network-based segmentation quality prediction methods: Reverse Classification Accuracy (RCA) \cite{valindria2017reverse} and QualityNet \cite{huang2016qualitynet}. As each of these approaches have their own architectural and optimization-based hyper-parameters, we have kept these the same as our technique, where applicable.

\subsection{Uncertainty Maps}
To compare the quality of the uncertainty maps from each of the different uncertainty (and confidence) estimation techniques, we visualize them in Figure \ref{fig:uncertainty_maps}. In cases where the predicted segmentation is very close to the ground truth segmentation, we find that all techniques act similarly, outputting a tight ring along the segmentation borders. This is what we would expect in such a situation. The more interesting observation is how the uncertainty maps react when the predicted segmentation is very poor. We find that in general, maximum softmax probabilty, MC-dropout, and LCE all display high uncertainty in regions that are segmented incorrectly. Conversely, HCNN usually outputs a small band of low uncertainty around its prediction, but does not highlight other areas that may be incorrect. This output is less useful for identifying failed segmentations, which agrees with our findings in \S \ref{section:segmentation_quality_prediction}.

\subsection{Segmentation Quality Prediction}
\label{section:segmentation_quality_prediction}
We use a variety of metrics to evaluate how well our models can predict the quality of segmentations: RMSE, detection error, AUROC, and AUPR; each of which is defined below.

\textbf{RMSE}: Measures Root Mean Squared Error, which is the difference between the predicted Jaccard index and the true Jaccard index. Predictions that are further from the true value are penalized more heavily in this metric. RMSE is calculated as $\sqrt{\frac{\sum_{t=1}^{n}({\hat{v}}_{t} - v_{t})^{2}}{n}}$ where $t$ indexes the test examples, and $n$ is the total number of test examples.

For practical applications (e.g.~human-in-the-loop) we may also want to measure how well our model can detect \emph{failed} segmentations. For the ISIC 2017 dataset, a Jaccard index of below 0.7 is considered to be a failed segmentation \cite{codella2017skin}. To evaluate how well our model can detect these failures, we threshold the true Jaccard index labels at 0.7 to obtain binary labels, which we can use to calculate detection error, AUROC, and AUPR.

\textbf{Detection Error}: Measures the minimum possible misclassification probability over all possible thresholds $\delta$ when detecting segmentation failures, as defined by $\min_{\delta}\left\{0.5 \, P_{\text{pass}}(f(x)\leq \delta) + 0.5 \, P_{\text{fail}}(f(x)> \delta) \right\}$. Here, we equally weight $P_{\text{pass}}$ and $P_{\text{fail}}$ as if they have the same probability of appearing in the test set.

\textbf{AUROC}: Measures the Area Under the Receiver Operating Characteristic curve. The Receiver Operating Characteristic (ROC) curve plots the relationship between true positive rate and false positive rate. The area under the ROC curve can be interpreted as the probability that a correctly segmented image will have a higher quality estimate than a failed segmentation.

\textbf{AUPR}: Measures the Area Under the Precision-Recall (AUPR) curve, which is calculated by plotting precision versus recall. In our tests, AUPR-Pass indicates that acceptable segmentations are used as the positive class, and AUPR-Fail indicates that failed segmentations are used as the positive class. We evaluate both metrics so that we can see if our model is biased towards either class.

% \begin{figure*}[!h]
% \centering
% \includegraphics[width=0.8\textwidth]{jaccard_estimation_scatter_plots_higher_res.pdf}
% \caption{Scatter plot of the predicted vs.~true Jaccard index from a segmentation quality estimation network trained on LCE. Here we annotate some points of interest with their associated image, ground truth, and the predicted segmentation. We note that most false positives (upper left corner) are caused by corrupted labels or lazy annotations.}
% \label{fig:annotated_jaccard_scatter}
% \end{figure*}

\begin{figure*}[!h]
\centering
\begin{minipage}{0.9\textwidth}
\includegraphics[width=\textwidth]{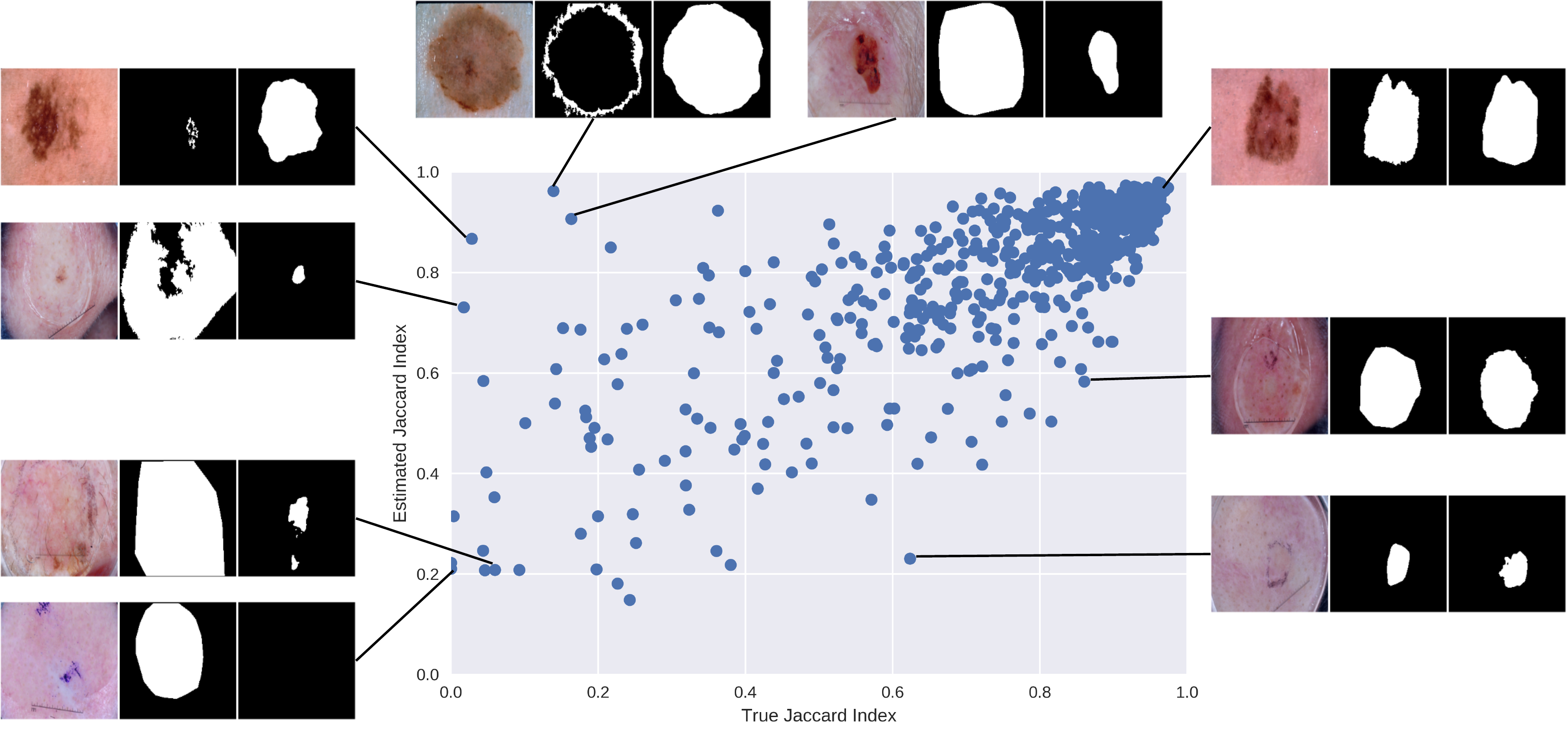}
\subcaption{LCE}
\label{fig:annotated_jaccard_scatter}
\end{minipage}
\begin{minipage}{0.45\textwidth}
 \includegraphics[width=\linewidth, trim={1cm 0cm 1cm 1cm}, clip]{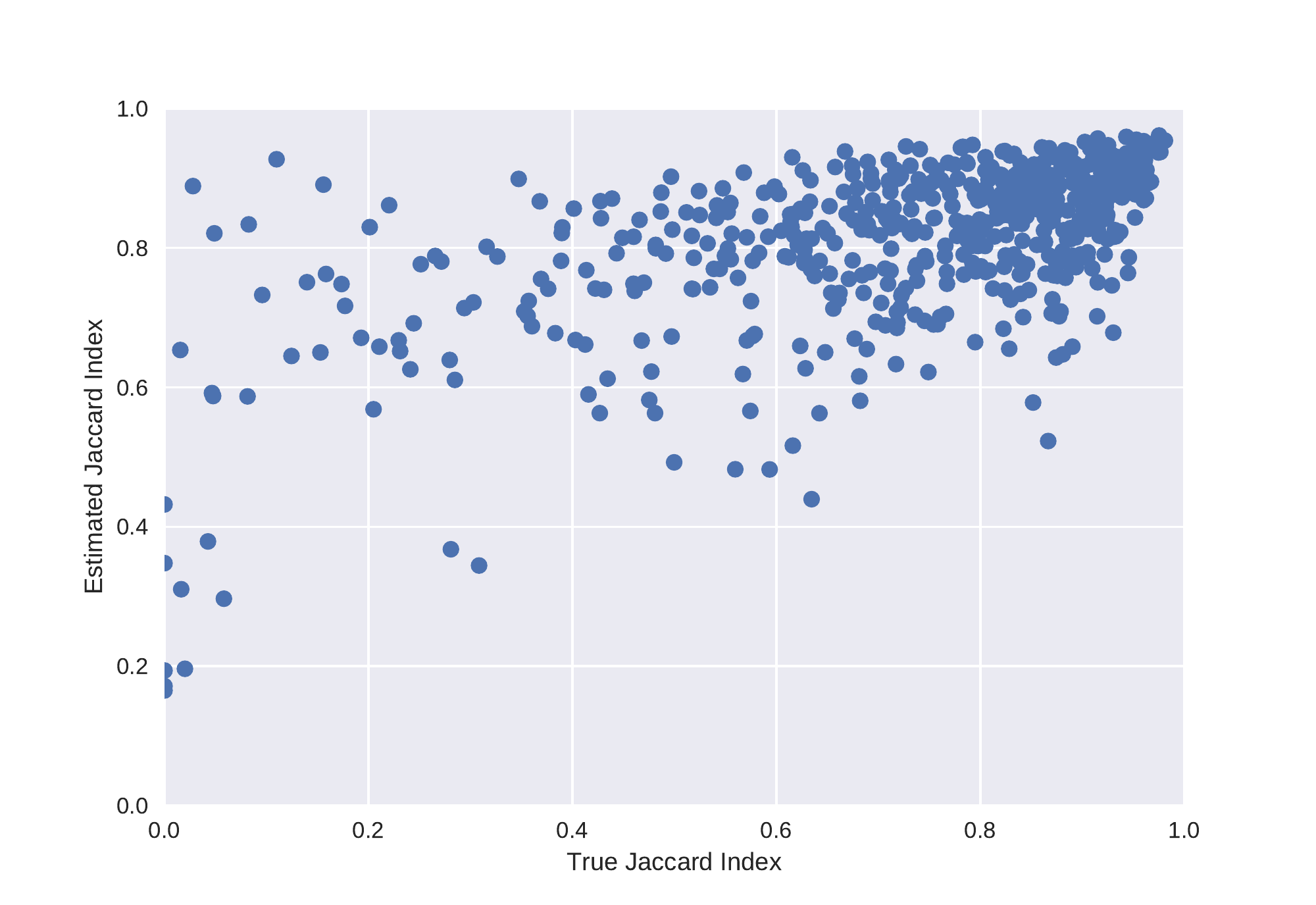}
 \subcaption{No uncertainty}
\end{minipage}
\begin{minipage}{0.45\textwidth}
 \includegraphics[width=\linewidth, trim={1cm 0cm 1cm 1cm}, clip]{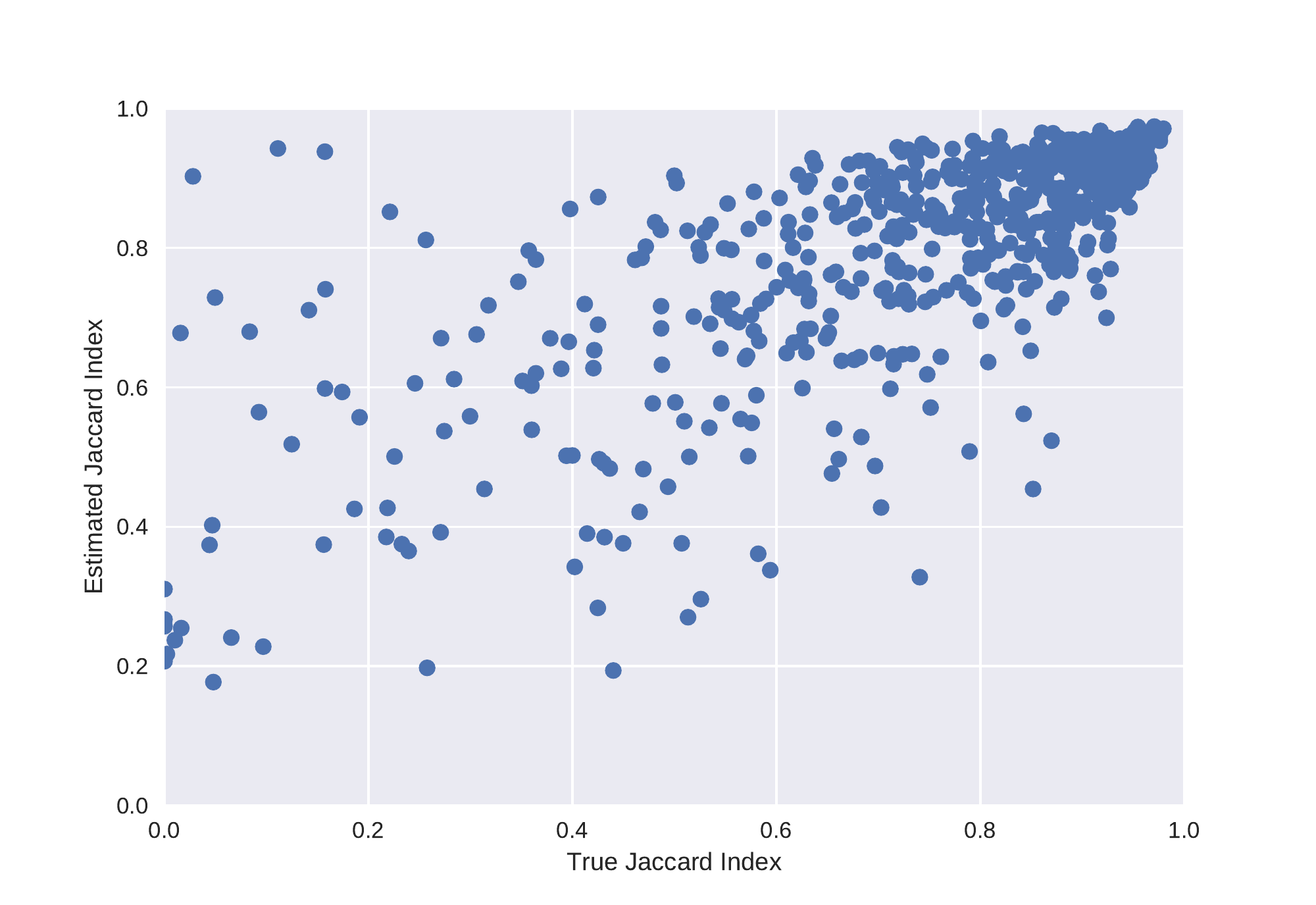}
 \subcaption{Max probability}
\end{minipage}\\
\begin{minipage}{0.45\textwidth}
 \includegraphics[width=\linewidth, trim={1cm 0cm 1cm 1cm}, clip]{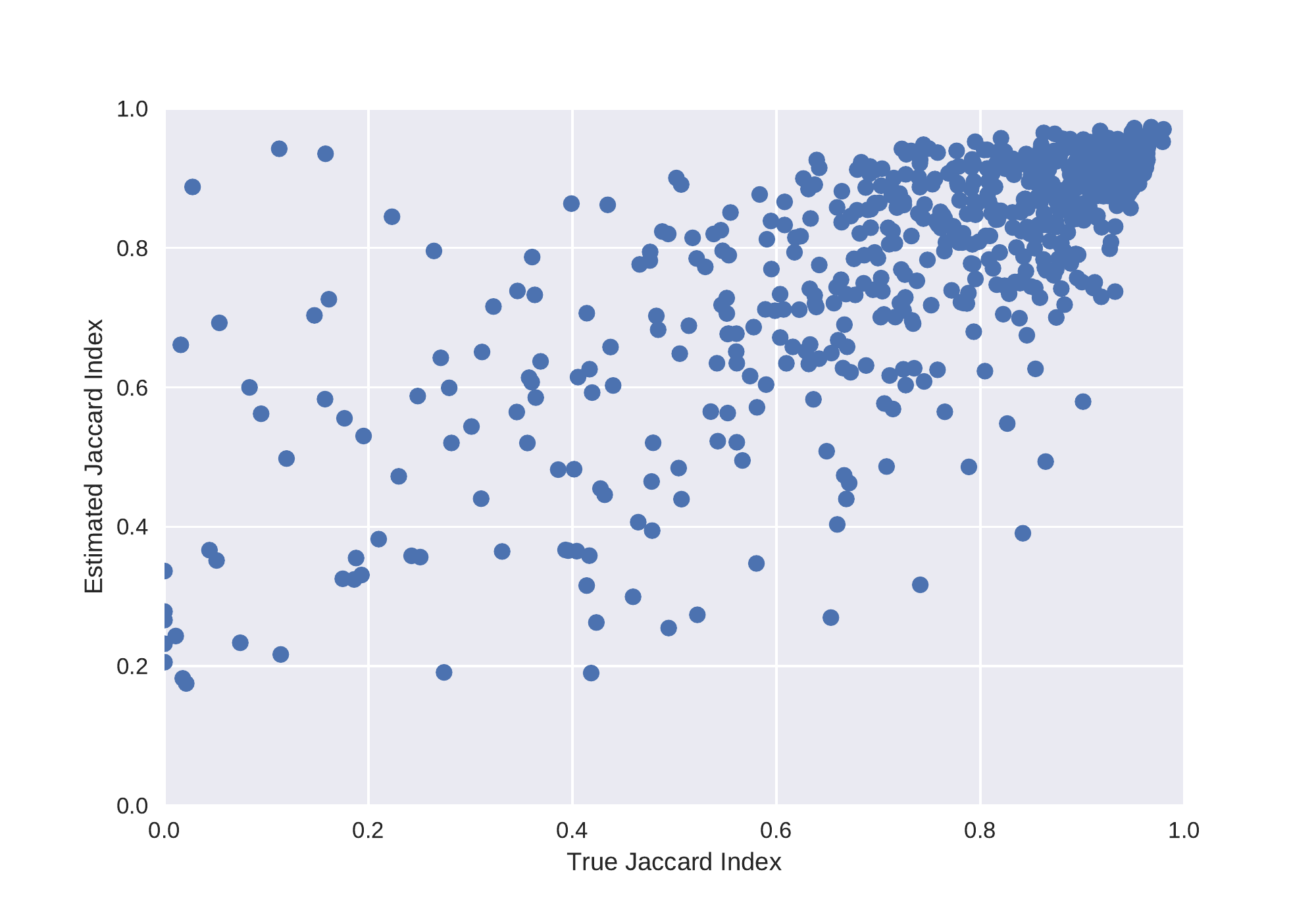}
 \subcaption{MC-dropout}
\end{minipage}
\begin{minipage}{0.45\textwidth}
 \includegraphics[width=\linewidth, trim={1cm 0cm 1cm 1cm}, clip]{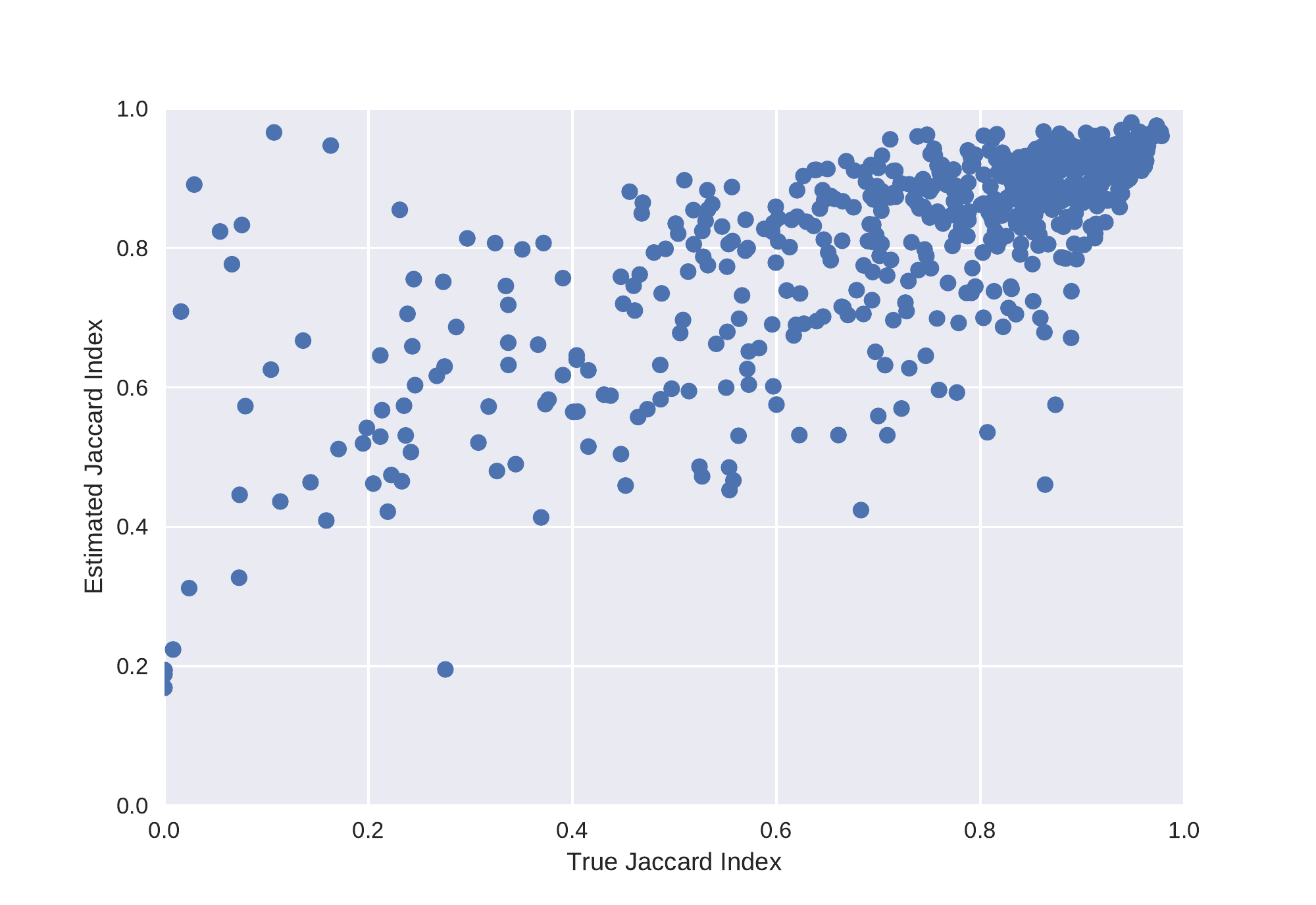}
 \subcaption{HCNN}
\end{minipage}
\caption{Scatter plots of true Jaccard index versus the estimated Jaccard index from segmentation quality estimation networks trained with different uncertainty estimate techniques. In a) we annotate some points of interest with their associated image, ground truth, and the predicted segmentation respectively. We note that most false positives (upper left corner) are caused by corrupted labels or lazy annotations.}
\label{fig:all_jaccard_scatter}
\end{figure*}

In Table \ref{table:seg_quality_prediction}, we present the results of our quantitative evaluation of different segmentation quality estimation methods. These are organized by two groupings: 1) recent baselines RCA and QualityNet, which leverage CNNs to predict segmentation quality and 2) variants of our two-stage CNN approach. The variants of our approach apply different quantitative measures of uncertainty, described above. We also include a method which does not explicitly generate uncertainty as an intermediate representation, i.e.~$g$ receives only $(x,\hat{y})$ as input rather than $(x,\hat{y},z)$. This is slightly different than QualityNet, in which $g$ receives $(x \odot{\hat{y}})$, where $\odot{}$ is an element-wise matrix multiplication.

We find that treating uncertainty explicitly improves performance significantly compared to the case with no uncertainty information, reducing RMSE by up to 0.03 points, and detection error by up to 8 percentage points. However, surprisingly the particular method by which uncertainty is captured does not have a large effect, at least in this setting. Maximum softmax probability, MC-dropout, and LCE all produce similar improvements in performance. It was expected that MC-dropout or LCE would outperform maximum softmax probability since they have been shown to surpass softmax probability in tasks such as out-of-distribution detection \cite{devries2018learning}, but this was not the case for this particular dataset. Of the uncertainty estimation techniques, HCNN improved performance the least; only 5 points of detection error and no improvement on RMSE. This agrees with \cite{kendall2017uncertainties}, which indicates that aleatoric uncertainty which HCNN aims to capture, is a poor choice for detecting model failures since it mainly models noise in the data itself.

We find that our implementation of QualityNet performs roughly equal to our \emph{no uncertainty} baseline, which is expected given how similar the implementations are. Unfortunately, RCA performs very poorly; only slightly better than random. This is likely because the algorithm was designed to work on datasets of registered images with very little variation between them, such as MRI scans of internal organs. In these datasets each of the objects to be segmented are extremely similar in shape and location. In contrast, the skin lesions from the ISIC 2017 dataset appear with a wide variety of colours, textures, shapes, sizes, and locations, making them very difficult for modern image registration techniques to succeed.
%We note that the authors of RCA found Single-Atlas to work better than CNN for their segmentation quality estimation task

In Figure \ref{fig:all_jaccard_scatter} we plot the true versus the predicted Jaccard index for each of the different uncertainty estimates we tested. We observe that max probability, MC-dropout, and LCE are all better at identifying poor quality segmentations (lower left corner) compared to HCNN or the no uncertainty baseline. Additionally, we note that the majority of false positives (poor quality segmentations that are rated highly) are caused by either corrupted labels or lazy annotations, as shown in Figure \ref{fig:annotated_jaccard_scatter}.

Interestingly, segmentation quality estimates rarely fall below 0.2 for any method. This is likely caused by the rarity of poor quality segmentations in the dataset used to train the segmentation quality estimation network, since it was trained on the same dataset as the original segmentation network. While it is probable that using a separate held-out dataset would result in a greater number of poor quality segmentation examples, and therefore better performance from the segmentation quality estimation network, we do not explore this option due to the small size of the ISIC 2017 dataset.

\section{Conclusion}
In this work, we investigated techniques which aid a human operator, such as a clinician, interact with a deep learning-based automated segmentation pipeline. We showed how uncertainty could be derived at the pixel- and image-level within a single end-to-end framework. We demonstrated our method qualitatively and quantitatively on the task of skin lesion segmentation. Though a neural network trained to predict segmentation quality has the capacity to measure uncertainty internally, we showed that making spatial uncertainty explicit aided in predicting a measure of segmentation quality, the Jaccard index. Moreover, we demonstrated that several recent methods for quantifying uncertainty worked well in this setting. In the future, we plan on extending our analysis to other medical segmentation problems, and even tasks outside segmentation that could benefit from a human-in-the-loop. We also used simple, standard losses for our segmentation model. Recent techniques that aim to optimize application-specific metrics like the Jaccard index would likely improve overall performance.

{\small
\bibliographystyle{ieee}
\bibliography{egbib}
}

\end{document}